\documentclass{esannV2}
\usepackage{graphicx}
\usepackage[latin1]{inputenc}
\usepackage{amssymb,amsmath,array}
\usepackage{subcaption}
\usepackage{amsthm}
\usepackage{url}
\newtheorem{definition}{Definition}

\begin{document}
\title{Federated Learning\\ Methods, Applications and Beyond}

\author{Moritz Heusinger$^1$, Christoph Raab$^1$, Fabrice Rossi$^2$ and Frank-Michael Schleif$^1$
%
%
\vspace{.3cm}\\
%
1- University of Applied Science W{\"u}rzburg-Schweinfurt - Department \\ of Computer Science, W{\"u}rzburg - Germany \\
2- CEREMADE, University Paris Dauphine PSL, France
}


\maketitle

\begin{abstract}
In recent years the applications of machine learning models have increased rapidly, due to the large amount of available data and technological progress. 
While some domains like web analysis can benefit from this with only minor restrictions, other fields like medicine with patient data are stronger
regulated. In particular \emph{data privacy} plays an important role as recently highlighted by the trustworthy AI initiative of the EU or general privacy regulations in legislation.  Another major challenge is, that the required training \emph{data is} often \emph{distributed} in terms of features or samples and unavailable for classical
batch learning approaches. In 2016 Google came up with a framework, called \emph{Federated Learning} to solve both of these problems. We provide a brief overview on existing Methods and Applications in the field of vertical and horizontal \emph{Federated Learning}, as well as \emph{Federated Transfer Learning}.

\end{abstract}

\section{Introduction}
Federated learning (FL) is a novel concept for learning distributed data, which was first introduced by Google \cite{konevcny2016federatedL,konevcny2016federatedO, McMahan2016FederatedLO} in 2016.

\begin{figure}[ht]
    \centering
    \includegraphics[width=0.95\textwidth]{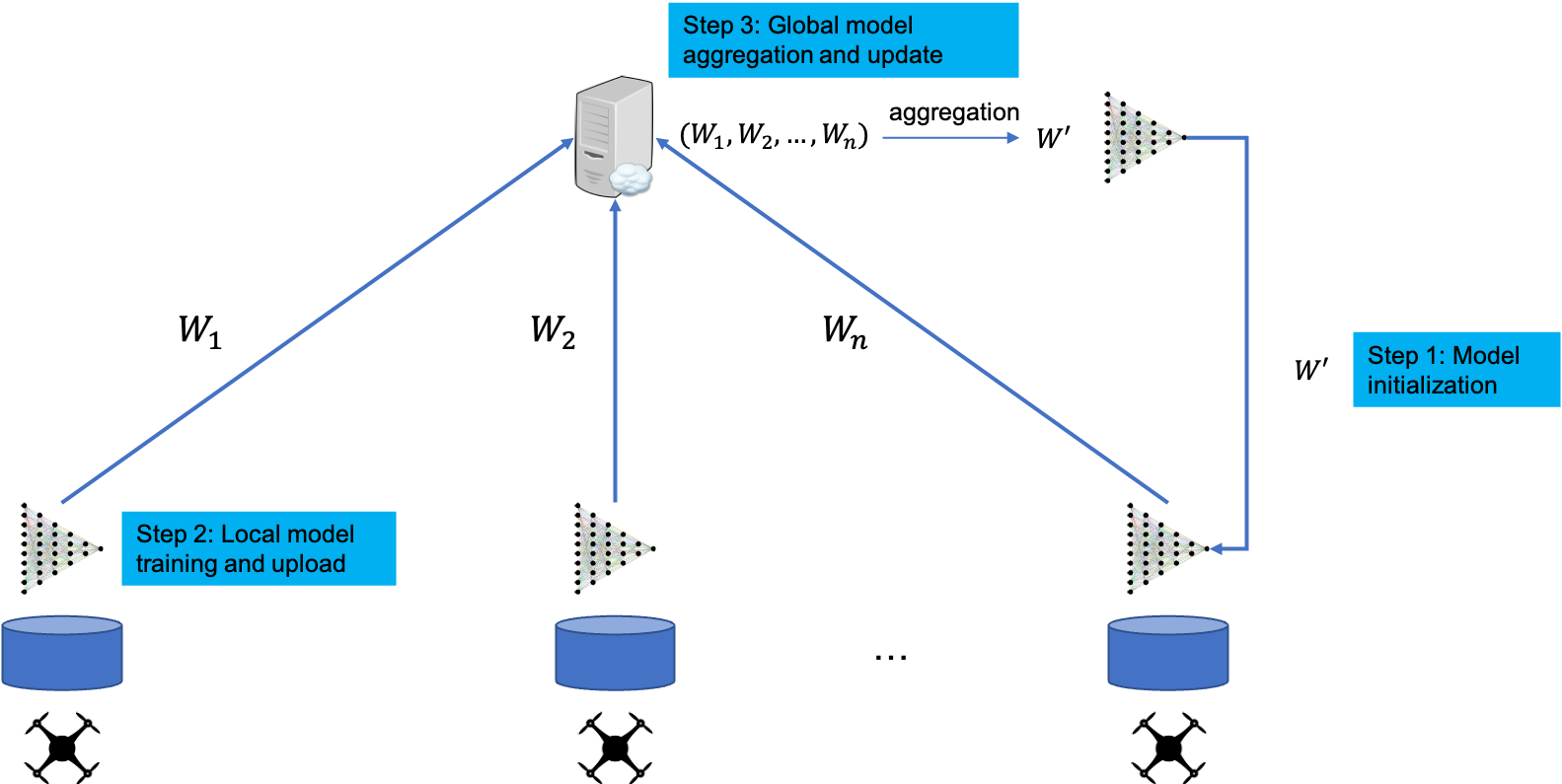}
    \caption{Diagram of FL. To preserve data privacy, local model gradients are only sent to one trusted server (or a primary coordinator) and not directly to other clients. The local instances are training their own model with the local data and after multiple iterations the gradient is sent to the primary coordinator. The primary coordinator aggregates all local gradients to a central global update, which is used to update the global model. Finally, the global model is sent back to the clients to replace their previous models.}
    \label{fig:fl-flow}
\end{figure}

\begin{definition}[Federated Learning]
Given a large number of $N$ clients and a particular data analysis task,
each client $C_i$ has its own data addressing the task, without direct
access to the other clients data.
The objective in FL is to learn a predictive model $M$ such that
the error on the objective function $E$ is minimized, in a distributed way. In particular
various data processing clients are involved. The communication
    takes place by a distributed protocol where in general a master is
identified to aggregate the prediction model.
    FL has three steps (1) an initial model is distributed to the
clients (2) the local model of $C_i$ is trained on its local data by
taking the model information of the master into account
    (3) the master aggregates the local models $M_i$ to the global model
$M$ and communicates the global model back to the clients.\label{def:fl}
The steps are visualized in Figure \ref{fig:fl-flow}.
\end{definition}

FL has gained substantial interest in the machine learning (ML) community with different frameworks implementing the main concept \cite{10.1145/3437963.3441702}, in particular \emph{flower} \cite{beutel2021flower} which is 
considered very mature. Applications of FL are more and more frequent \cite{BRISIMI201859,LI2020106854,YAZDINEJAD2021102574,FML}. The research field is also very active with new  communication protocols \cite{konevcny2016federatedL}, encryption concepts \cite{MOTHUKURI2021619}
and particular optimization algorithms \cite{konevcny2016federatedO,MODES,8744465}. 

FL has numerous advantages over classical ML. While in classical machine learning training data are submitted to a central instance and a model is learned in batch processing, FL shifts the actual learning to the data source.
This allows one to employ the power of distributed client machines, keeps the user data private, and permits to use information that is otherwise inaccessible and spread over different clients.

In this view FL perfectly aligns with recent trends on machine learning on large community data \cite{YANG2021106946} and the increasing set of constraints due to privacy regulations like the GDPR\footnote{\scriptsize\url{https://gdprinfo.eu/}},
trustworthy AI\footnote{\scriptsize\url{https://digital-strategy.ec.europa.eu/en/library/ethics-guidelines-trustworthy-ai}} but also objectives covered in various AI manifests\footnote{\scriptsize\url{https://nouvelles.umontreal.ca/en/article/2018/12/04/developing-ai-in-a-responsible-way/}}.

Broadly speaking, Federated Learning methods belong to three different categories, which depend on how comparable are the local data of the clients, as detailed in Sec. \ref{sec:fl-types}.

\section{Types of federated learning}\label{sec:fl-types}
We denote the data held by client $C_i$ as $\mathcal{D}_i$. In FL a data set consists of the features $\mathbf{X}$, the sample id space $\mathbf{i}$ and an optional label space $\mathbf{y}$. The sample space, as well as the feature space, may not be identical over different data owners. By this characteristic, FL is categorized into horizontal FL (HFL), vertical FL (VFL), and Federated Transfer Learning (FTL), as detailed below.


\begin{figure}[ht]
    \centering
    \begin{subfigure}[]{.2\linewidth}
    \includegraphics[width=1\textwidth]{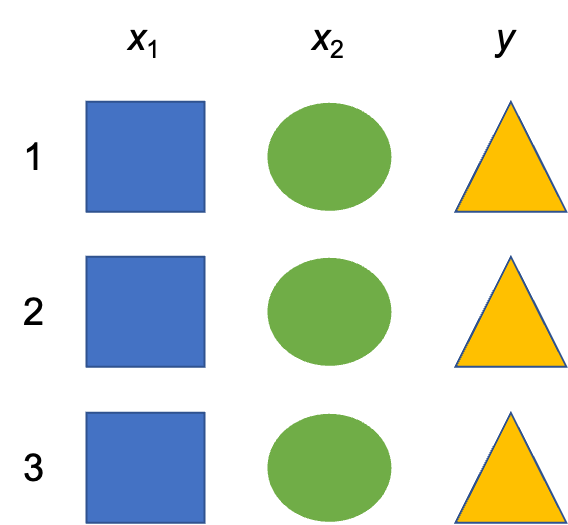}
    \subcaption{Client A}
    \end{subfigure}\hspace{20mm}
    \begin{subfigure}[]{.2\linewidth}
    \includegraphics[width=1\textwidth]{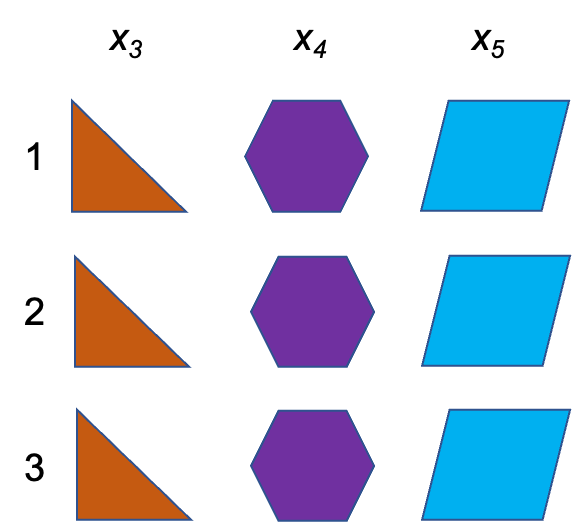}
    \subcaption{Client B}
    \end{subfigure}
    \caption{Vertical Federated Learning (inspired by \cite{chen2020})}
    \label{fig:vertical-fl}
    
    \begin{subfigure}[]{.2\linewidth}
    \includegraphics[width=1\textwidth]{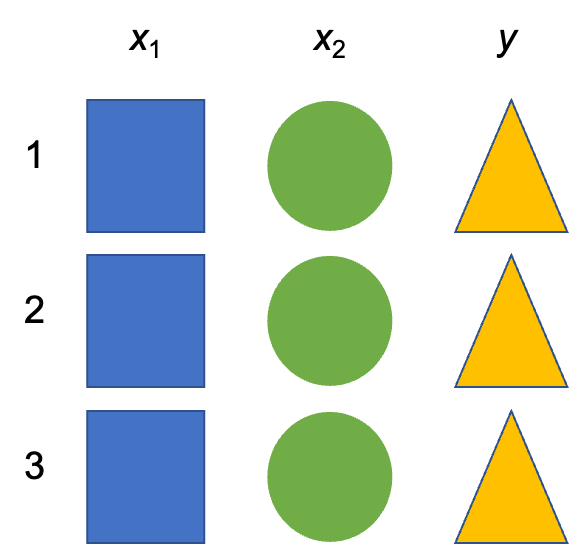}
    \subcaption{Client A}
    \end{subfigure}\hspace{20mm}
    \begin{subfigure}[]{.2\linewidth}
    \includegraphics[width=1\textwidth]{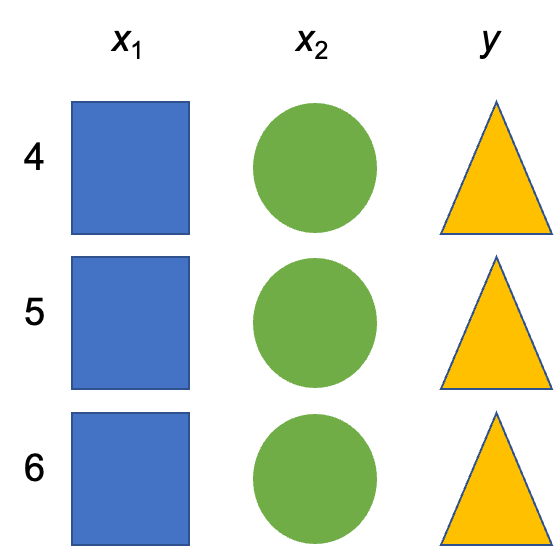}
    \subcaption{Client B}
    \end{subfigure}
    \caption{Horizontal Federated Learning (inspired by \cite{chen2020})}
    \label{fig:horizontal-fl}
    
    \begin{subfigure}[]{.2\linewidth}
    \includegraphics[width=1\textwidth]{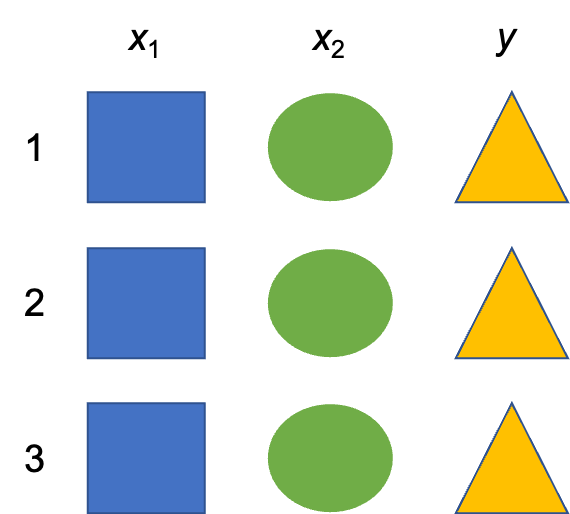}
    \subcaption{Client A}
    \end{subfigure}\hspace{20mm}
    \begin{subfigure}[]{.2\linewidth}
    \includegraphics[width=1\textwidth]{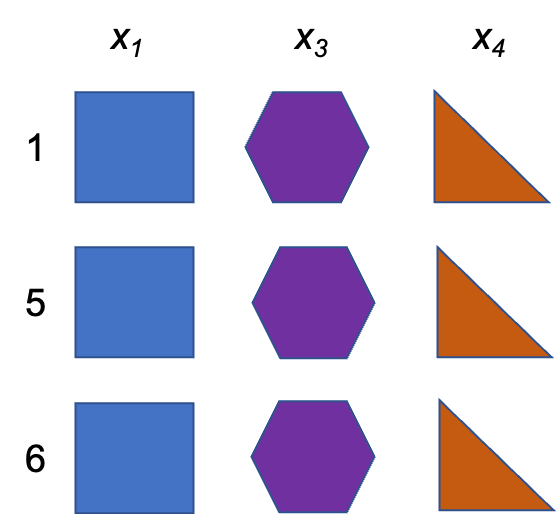}
    \subcaption{Client B}
    \end{subfigure}
    \caption{Federated Transfer Learning (inspired by \cite{chen2020})}
    \label{fig:transfer-fl}
\end{figure}

\subsection{Horizontal FL}\label{sec:hfl}
The scenario of HFL is depicted in Figure \ref{fig:horizontal-fl} and shows that the users are split across various clients, while the feature space is always the same. 
A typical scenario is given by a global business model which is implemented on smartphones or other IoT devices. Thereby the user generates the same data, like interaction events or shopping activities but may be located in very different geographical regions. HFL is beneficial by calculating a model which employs the information from a 
large (distributed) user group instead of focusing on a centralized approach with a rather limited amount of training data. In HFL it is common to calculate and upload local gradients calculated from the objective function, which are aggregated by a central master client. The data transmissions can be encrypted to improve the level of privacy using homomorphic encryption \cite{fi13040094}, differential privacy \cite{RODRIGUEZBARROSO2020270} or secure aggregation as discussed in Sec. \ref{sec:privacy}.

\subsection{Vertical FL}\label{sec:vfl}
Vertical federated learning considers the case where we have multiple different feature sets on a common basis of users. A typical case for this scenario (depicted in Figure \ref{fig:vertical-fl})
is an analysis task where the user is using at least two ways of interaction and is generating data on two channels, like an online store and in a brick and mortar business. 

\subsection{Transfer learning}\label{sec:ftl}
As described above, in FL, it is generally assumed that every user provides a sufficient set of labeled data for a model to learn a specified task. In general, it is also assumed that the analyzed data are given in common feature spaces, although potentially split across various clients \cite{FML}. If the first assumption does not hold, the model cannot capture the whole data set characteristic leading to poor prediction performance. Further, the provided gradients are biased towards the present data characteristics, which eventually deteriorates the global learning model. If the second assumption does not hold, the entities and the corresponding feature spaces are disjoint. This prevents the model from discovering the feature space characteristics, and therefore, the FL system cannot extract useful information for the specific user model \cite{FML}. 

In the cases just described, methods of Federated Transfer Learning need to be applied. The FTL setup is illustrated in Figure \ref{fig:transfer-fl}. FTL can be seen as the cross-section of horizontal and vertical transfer learning. In general, Transfer Learning is an algorithmic technique to improve a model trained on one data set, by using related information from another data set. The just mentioned data sets are usually denoted as source and target domain. FTL helps to improve the model of user $U_i$ on its data by using data or model information from one (or more) users $U_j$, where $i\neq j$. Hence, the learning environment of $U_i$ is called target, and $U_j$ is source \cite{Liu2020}. Note, that there can be small intersections regarding feature space or sample space as displayed in Figure \ref{fig:transfer-fl}. A model improvement on the target by means of FTL is achieved by creating a shared representation keeping the federation of data. For example via manifold alignment \cite{9175344} or domain adversarial learning \cite{PengHZS20}. Alternatively, FTL methods improve the target model leveraging the source model by instance reweighting \cite{9005992} or by receiving gradients from source \cite{Liu2020}. Note that some approaches also consider different data distributions between source and target \cite{9005992}.

\section{Frameworks and Algorithms}\label{sec:frameworks}
In FL the communication architecture and protocols are of particular importance. FL has to deal with non i.i.d. data \cite{8889996} and due to the FL learning concept substantial communication costs can occur. Considering the typical use cases of FL in the field of mobile devices, IoT \cite{9253545}, unmanned vehicles \cite{YAZDINEJAD2021102574} or large distributed server systems additional compression techniques, averaging strategies, and sparsity constraints are applied to obtain real time-efficient systems \cite{Chene2024789118,konevcny2016federatedL,konevcny2016federatedO}.
One can also employ quantization approaches \cite{TONELLOTTO2021417} or information about the transmitted data to control the communication load \cite{9303442}.

Only recently some FL frameworks have been proposed which simplify the implementation of own FL models. In particular, for HFL the BlockFL framework
was proposed in \cite{kim2019blockchained} which makes use of a blockchain during the updates of the model parameters. A framework for VFL is provided
by SecureBoost \cite{9440789}. And in the context of Federated Transfer Learning the framework in \cite{9076003} is suggested. The framework \emph{Flower} \cite{beutel2021flower}
scales well to a large number of clients.

The machine learning community has also recently started to design dedicated learning algorithms for FL. One example is given in \cite{MODES} where hyperparameter
learning on distributed systems is considered. Also information theoretic strategies have been proposed \cite{9272656}, matrix factorization techniques \cite{YANG2021106946},
spectral clustering \cite{9252122}, particular designed gradient descend techniques \cite{9003425,Fernandes_ESANN21} or multi-objective solvers \cite{8744465}. Furthermore, the Learning Vector Quantization (LVQ) concept has been adopted, to fit in a VFL environment by training separate LVQ models locally and using the relevance matrix to update a global model \cite{Brinkrolf_ESANN21}.

\section{Privacy methods}\label{sec:privacy}
The key element of all FL approaches is to keep the data on the user side and in particular to avoid any disclosures. Three techniques
are most common to ensure this goal and are frequently combined: 
\begin{enumerate}
    \item homomorphic encryption 
    \item differential privacy 
    \item (secure) model aggregation
\end{enumerate}
The most common is model aggregation which trains the global model by summarizing the model parameters from all clients to avoid disclosure of original data.
Many optimization concepts which are roughly based on a kind of iterative stochastic gradient descent on $E$ perform a natural averaging of various update steps which is also used in batch online learning approaches \cite{bishop1995neural}. A prominent approach for deep learning falling into this category is given in \cite{McMahan2016FederatedLO}.
An alternative view is to train local models in a multi-task setting which are subsequently combined as shown in \cite{DBLP:conf/icml/YurochkinAGGHK19}. The various
local model parameters can also be safely transferred in an aggregated form by blockchain techniques as shown in \cite{kim2019blockchained}. One may also directly provide privacy-preserving data representations as shown for kernels in \cite{Polato_ESANN21}.

Homomorphic encryption allows to apply calculations on encrypted data without the need of decoding. One approach following this idea is additive homomorphism
\cite{hardy2017private}.

Differential privacy as detailed in \cite{DBLP:conf/icalp/Dwork06} is a technique to limit information disclosure during learning. Thereby the training procedure is
designed such that small modifications of the training database have no substantial impact on the model outcome. The attacker can not obtain accurate
individual information, but only a controlled piece of information, which still obeys privacy constraints. The particular strategies to implement this
concept can be very simple by adding some noise contributions to the output during training or by more complex compression techniques as shown in \cite{9069945}.

In \cite{Polato_ESANN21} a privacy-preserving method to compute dot-product kernels in VFL is proposed. The technique uses multi-party computation to provide theoretical guarantees on security and privacy.
For a more detailed analysis, we refer to \cite{MOTHUKURI2021619}.
\section{Applications}
FL can train a united model on data from distributed sources while preserving data privacy and security and thus can play an important role in many industrial sectors, like sales, health, insurance, and others. In general, in every sector, where data cannot be directly aggregated due to privacy protection, data security, or even property rights \cite{FML}.

FL has been successfully used to improve the quality of keyboard search suggestions on the Google Gboard \cite{gboard2018}. Gboard is a virtual keyboard for mobile devices and has several features, like auto-completion and next-word prediction. The application needs to protect the privacy of users and make latency-free predictions. To avoid high data usage and battery consumption due to the FL optimization, the authors had to design mechanics to only send data to a centralized server, if the device is inside a wireless network and actively charging. The server provides every client with a training task as soon as enough clients are connected. In a related context, FL has also been used to perform mobile keyboard predictions \cite{leroy2019federated}.

Recently, a project has been started, which uses FL to deploy an auction for intralogistic autonomous drone transportation\footnote{mFUND program of the BMVI, project FlowPro, grant number 19F2128B}. The goal of the project is, to create a system where drone owners can bid on transport jobs and execute them in case of winning. The bidding model is based on FL, which enables the knowledge incorporation of every drone in the global model of the system. The distribution of the global model also leads to the fact, that drones with fewer executed jobs will have the same chance to win an auction as more experienced drones.

In \cite{Pellegrini_ESANN21}, a hybrid continual learning strategy is used to address the real-world constraints like computational and memory limits in a real-time on-device personalization task, running on a native Android application.

Other applications of FL include ranking browser history suggestions based on user-interactions \cite{hartmann2019federated}, visual object detection \cite{liu2020fedvision}, patient clustering to predict hospital stay time as well as mortality \cite{HUANG2019103291}, drug discovery \cite{Xiong2020FacingSA,Chen2020.02.27.950592} and brain tumor segmentation \cite{Li2019AbnormalCB,flBrainTum2018,brainTum2019}.
The application fields of FL are quickly increasing and are a promising research direction of ML. For a more comprehensive review, we refer to \cite{LI2020106854}.

\section{Conclusions}
In this tutorial, we briefly discussed the evolving field of federated learning and outlined recent achievements and approaches. A detailed analysis of recent trends and problems in FL is also provided in \cite{AdvProbFL,SurveyFL,9153560,9084352}. Due to additional privacy constraints in ML and a variety of distributed user groups of ML methods, it can be expected that FL will become even more important in the future

\begin{footnotesize}

{\flushleft{\bf Acknowledgment}}
{\flushleft
MH and CR are thankful for support in the FuE program Informations- und Kommunikationstechnik of the StMWi, project OBerA, grant number IUK-1709-0011// IUK530/010 and for support in the mFUND program of the BMVI, project FlowPro, grant number 19F2128B.}



\bibliographystyle{unsrt}
\bibliography{library,session}

\end{footnotesize}


\end{document}